\documentclass[conference]{IEEEtran}
\IEEEoverridecommandlockouts
\usepackage{cite}
\usepackage{url}
\usepackage{amsmath,amssymb,amsfonts}
\usepackage{graphicx}
\usepackage{textcomp}
\usepackage{xcolor}
\usepackage{hyperref}
\usepackage{cite}
\usepackage{comment}
\usepackage{amsmath,amssymb,amsfonts}
\usepackage{graphicx}
\usepackage{textcomp}
\usepackage{balance}
\usepackage{tabularx}
\usepackage{makecell}
\usepackage{multirow}
\usepackage{algorithm}
\usepackage{algpseudocode}

\usepackage{amsmath}
\usepackage{subcaption} 
\usepackage[table,xcdraw]{xcolor}
\def\BibTeX{{\rm B\kern-.05em{\sc i\kern-.025em b}\kern-.08em
    T\kern-.1667em\lower.7ex\hbox{E}\kern-.125emX}}

\begin{document}

\title{EROAS: 3D Efficient Reactive Obstacle Avoidance System for Autonomous Underwater Vehicles using 2.5D Forward-Looking Sonar\\
}


\author{\IEEEauthorblockN{Pruthviraj Mane}
\IEEEauthorblockA{\textit{Department of Aerospace Eng.} \\
\textit{Indian Institute of Science}\\
Bangalore, India \\
pruthvirajm@iisc.ac.in}
\and
\IEEEauthorblockN{Allen Jacob George}
\IEEEauthorblockA{\textit{Department of Electrical and Electronics Eng.} \\
\textit{Birla Institute of Technology and Science}\\
Hyderabad, India \\
allenjacobgeorge@gmail.com}
\and
\IEEEauthorblockN{Rajini Makam}
\IEEEauthorblockA{\textit{Department of Aerospace Eng.} \\
\textit{Indian Institute of Science}\\
Bangalore, India\\
rajinimakam@iisc.ac.in}
\and 
\IEEEauthorblockN{Subhash Gurikar}
\IEEEauthorblockA{\textit{Department of Aerospace Eng.} \\
\textit{Indian Institute of Science}\\
Bangalore, India \\
subhashgurikar7@gmail.com}
\and
\IEEEauthorblockN{Rudrashis Majumder}
\IEEEauthorblockA{\textit{Department of Computer Science and Eng.} \\
\textit{Shiv Nadar University}\\
Chennai, India \\
rudrashism@snuchennai.edu.in}
\and
\IEEEauthorblockN{Suresh Sundaram}
\IEEEauthorblockA{\textit{Department of Aerospace Eng.} \\
\textit{Indian Institute of Science}\\
Bangalore, India \\
vssuresh@iisc.ac.in}
}

\maketitle

%

\begin{abstract}

Autonomous Underwater Vehicles (AUVs) have advanced significantly in obstacle detection and path planning through sonar, cameras, and learning-based methods. However, safe and efficient navigation in cluttered environments remains challenging due to partial observability, turbidity, the limited field-of-view of forward-looking sonar (FLS), and occlusions that obscure obstacle geometry. To address these issues, we propose the Efficient Reactive Obstacle Avoidance Strategy (EROAS), a lightweight framework that augments a standard 2D FLS with a pivoting mechanism, effectively transforming it into a cost-efficient \emph{2.5D sonar}. This design provides vertical information on demand, extending situational awareness while minimizing computational overhead. EROAS integrates three complementary modules: first, Sonar Profile-guided Directional Decision Control (SPD2C) for rapid gap detection and generation of reference commands in both horizontal and vertical planes. Secondly, the Spatial Context Generator (SCG), which maintains a short-term obstacle memory of the past to mitigate partial observability, and finally, a Spatio-Temporal Control Barrier Function (ST-CBF) that enforces forward-invariance of safety constraints by filtering nominal references. Together, these components enable robust, reactive avoidance of obstacles in uncertain and cluttered 3D underwater settings. Simulation and hardware-in-the-loop (HIL) experiments validate the efficacy of the proposed EROAS algorithm, demonstrating improved trajectory efficiency, reduced travel time, and enhanced safety compared to conventional methods such as the Dynamic Window Approach (DWA) and Artificial Potential Fields (APF). \url{https://github.com/AIRLabIISc/EROAS}

\end{abstract}


\section{Introduction}
Autonomous Underwater Vehicles (AUVs) are becoming increasingly vital for applications such as marine exploration, environmental monitoring, defense, and offshore infrastructure inspection \cite{ Roper2021, zhang2023}. To navigate in complex underwater environments effectively, AUVs must address challenges such as limited visibility, unpredictable currents, and unfamiliar terrains. Efficient and safe navigation is crucial for mission success, since collisions with unforeseen obstacles can damage the vehicle, or endanger protected objects, potentially resulting in mission failures or expensive repairs. The absence of GPS signals further complicates the localization process. Consequently, incorporating advanced obstacle detection and avoidance algorithms into AUVs is essential to improve their operational success, especially in areas where pre-mapped data are unavailable \cite{xanthidis2020}.

The AUVs employ various sensors for obstacle detection and avoidance. Optical sensors, such as cameras, offer high-resolution images of excellent detail in clear water, but their performance degrades significantly in turbid or poorly lit environments \cite{Alamdari2021TASE}. In contrast, 2D sonar systems perform well in low-visibility conditions by providing distance and depth information in a single plane. However, their effectiveness diminishes in complex three-dimensional environments due to their lack of vertical spatial awareness, which is crucial for accurate navigation and obstacle avoidance \cite{Morency2022IROS}. 

To address the need for three-dimensional situational awareness, full 3D sonar systems have been developed, offering detailed spatial data for accurate obstacle mapping. These systems provide information about depth and width to enable 3D navigation. However, the trade-off comes in the form of substantial computation and energy requirements. Processing large volumes of data produced by 3D sonar systems can overwhelm the AUV's onboard resources, making certain missions impractical. Recent alternatives with end-to-end learning with multi-sensor transformers for autonomous avoidance \cite{lin2023transformer} and cooperative stochastic Model Predictive Control (MPC) for ROV-AUV interaction \cite{2021MPC} show promise but they require substantial real world data and enhanced computation, thereby complicating the deployment on edge devices. 

To address these challenges in 3D underwater navigation, this paper introduces a novel obstacle avoidance method, EROAS. The main contributions of this work can be summarized as follows. First, we address the inherent 
constraints of underwater environments, where navigation is challenged by partial observability, 
limited field-of-view of FLS, turbidity, and occlusions that obscure 
obstacle geometry and distribution. To overcome these issues, we propose the EROAS, a modular framework that couples three novel components. 
(i) The SPD2C provides a fast and 
reactive mechanism to identify feasible gaps and generate nominal reference commands in both 
horizontal and vertical planes. (ii) The SCG enhances situational awareness by retaining short-term memory of past obstacles, unlike conventional methods that rely only on current data, ensuring safety under partial observability when obstacles leave the sonar FOV.
(iii) The ST-CBF enforces safety-critical 
constraints by filtering the nominal references generated by SPD2C and SCG, guaranteeing 
forward-invariance of a safe set over time. Together, these elements enable EROAS to achieve 
reactive, resource-efficient, and safe navigation in cluttered and uncertain 3D underwater 
environments. The framework is validated in both simulation and Hardware-In-The-Loop (HITL) 
experiments, demonstrating improved trajectory efficiency and safety compared to existing 
methods such as Dynamic Window Approach (DWA) and Artificial Potential Fields (APF).

The paper is organized as follows: Section \ref{related_works} reviews existing obstacle avoidance approaches for underwater environments. Section \ref{PD} introduces the problem definition, and Section \ref{schematic} outlines the schematic of the proposed framework. Section \ref{eroas} presents the EROAS algorithm in detail, including the integration of SPD2C, SCG, and ST-CBF. Section \ref{result} demonstrates the performance of EROAS in cluttered 3D environments. Finally, Section \ref{conclusion} provides concluding remarks and discusses future research directions.

\section{Related Works} \label{related_works}

Obstacle avoidance and navigation in autonomous underwater vehicles (AUVs) have advanced from 2D control systems to sophisticated 3D systems using sonar, cameras, and sensor fusion. Early AUV navigation relied on 2D methods with nonlinear Lyapunov-based controllers for stability \cite{khalaji2020OceanE}. Sampling-based techniques, such as those developed by \cite{hernandez2016IROS} and \cite{dai2024_diha}, facilitated rapid path generation and real-time obstacle avoidance. However, these methods were limited to 2D environments, restricting their effectiveness in handling vertical motion and 3D obstacle avoidance.
\subsection{Vision and Sonar Based Navigation}

Vision-based navigation in AUVs was advanced through imitation learning methods such as UIVNAV, enabling data gathering, obstacle avoidance, and navigation without localization in various environments \cite{lin2024ICRA}. \cite{perez2016Sensors} developed robust control systems to maintain reliability in murky conditions. To improve obstacle avoidance, forward-looking sonar (FLS) has been widely used and remains common on AUVs operating in low-visibility conditions. Recent advancements have integrated FLS with profiling sonar (PS) to enhance 3D mapping and vertical accuracy \cite{joe2021AR}. 

Sensor fusion techniques, such as the transformer-based dual-channel self-attention architecture, have refined collision avoidance by combining sonar and non-sonar data for real-time decision-making \cite{lin2023transformer}. However, such end-to-end models require large real-world multi-sensor co-registered datasets and higher compute budgets, which complicates training and edge deployment. Additionally, methods like the Intelligent Vector Field Histogram (IVFH) use a multi-beam FLS to optimize heading and pitch for efficient collision avoidance \cite{zhang2022IVFH}. Recent innovations include using sonar for contour tracking in underactuated vehicles to improve navigation accuracy along unknown paths \cite{Yan2024OceanE}. Deep learning techniques have also enhanced real-time obstacle avoidance, with end-to-end neural networks based on convolutional gated recurrent units (CGRUs) integrating static and dynamic feature extraction \cite{lin2023CGRU}. Recently, the authors in \cite{Noda2025} presented a particle filter–based navigation framework with a 3D observation model. However, collision avoidance, which is critical for ensuring safe navigation, was not addressed.

\subsection{3D Navigation and Deep Learning Approaches}
The evolution toward 3D navigation has driven the exploration of advanced sonar techniques. Dense 3D reconstructions using fused orthogonal sonar images have significantly improved target positioning and obstacle detection \cite{liu2024_orthosonar}. High-precision underwater 3D mapping has been achieved with imaging sonar, although computational challenges persist, as systems often process only a fraction of sonar frames, limiting real-time performance \cite{kim2021_3dmap}. Approaches like OptD have been developed to reduce computational load during 3D multibeam sonar data processing, allowing for faster map generation with minimal accuracy loss \cite{stateczny2019}. Despite these advancements, many of these techniques are not yet suitable for real-time obstacle avoidance due to their inherent processing time requirements.

Although significant progress has been made in 3D mapping, most systems are designed for environment reconstruction rather than real-time obstacle avoidance. Emerging 3D obstacle avoidance algorithms for AUVs utilize sonar data to navigate complex underwater environments by adjusting the AUV's heading to avoid detected obstacles, ensuring safe navigation \cite{cai2020_3d_obs}. For instance, a system employs a "vision cone" for safe navigation around obstacles, though it assumes simplified obstacle shapes and operates within a limited speed envelope \cite{wiig2020_3d_jfr}. Additionally, deep reinforcement learning (DRL) approaches have been explored for 3D path following and obstacle avoidance \cite{havenstrom2021_drl_3d}. Although effective in simulations, DRL based methods face challenges such as lack of formal safety guarantees, extensive training requirements, and difficulties handling complex dynamic environments \cite{bar2004estimation}.
\subsection{Safe Navigation With Control Barrier Function}
Control barrier function (CBF) \cite{ames2019control} is a mathematical concept that guarantees the safety of autonomous vehicles by restricting their states into safe sets. The paper \cite{deng2020collision} presents adaptive cruise control of AUVs where the vehicles should follow a desired trajectory satisfying constraints specified by a control barrier function to avoid collision with obstacles. In \cite{ozkahraman2020combining}, a multi-AUV coverage mission is performed with CBF as the safety constraint. To ensure the safety of higher relative degree models of AUVs, high-order control barrier functions (HOCBF) are used \cite{ wang2023safety}. 
Complementary to CBF-based safety, stochastic MPC (SMPC) has been used for cooperative collision avoidance between ROVs and AUVs, handling motion uncertainty and intent \cite{2021MPC}. Though effective in simulation, these methods are computationally intensive because of higher online optimization cost and rely on accurate models, limiting real-time deployment. An MPC-CBF combination for safe navigation in cluttered environment is dealt in \cite{Makam2025}. The vehicle dynamics is 3-DoF depth subsystem. Further, an hybrid-RRT$^*$ with CBF is presented for underwater environment \cite{Majumder2025}. 

\section{EROAS} \label{methodology}
This section introduces the problem formulation, presents the schematic of EROAS, and then provides a detailed explanation of the algorithm, including the integration of SPD2C, SCG, and ST-CBF, in the subsequent subsections.

\subsection{Problem Definition}\label{PD}

Safe autonomous navigation of AUVs in underwater terrain is a critical challenge due to the uncertain and unstructured environment that is most likely to be cluttered by obstacles of various shapes and sizes. To address the challenge, we introduce a novel approach called EROAS, where the objective is to generate safe reference commands $V^S_{\mathcal{R}}$ and $r_{\mathcal{R}}$, corresponding to the speed and yaw rate of the vehicle, respectively. The proposed EROAS algorithm considers constraints of (i) vehicle motion, (ii) partial observability due the FLS field of view (FOV), and (iii) safety issues due to uncertain underwater terrain. 

\subsubsection{Vehicle Model}
In our problem, the vehicle in consideration is based on the ROS-Gazebo-based DAVE simulator \cite{zhang2022dave}. The simulator features a hovering AUV named REXROV2 with six degrees of freedom (DOF) \cite{berg2012rexrov}, comprising three translational and three angular velocities  $\boldsymbol{\nu} = [v_x \quad v_y \quad v_z \quad p \quad q \quad r]^T$ and the control inputs are the six thrusters. The kinematics of REXROV2 is given by,
\begin{equation}\label{model}
    \dot{\boldsymbol{\eta}} = \boldsymbol{J}(\boldsymbol{\eta}) \, \boldsymbol{\nu},
\end{equation}
where, pose (i.e., position and orientation) $\boldsymbol{\eta} $ = $[\mathbf{p}_v \quad \Psi]$ with $\mathbf{p}_v =[x_v \quad y_v \quad z_v]^T$ and $\Psi =[\phi \quad \theta \quad \psi] ^ T$. Also,
\begin{equation}\label{jacobian}
\boldsymbol{J}(\boldsymbol{\eta}) = 
\begin{bmatrix}
R(\phi, \theta, \psi) & \mathbf{0}_{3 \times 3} \\
\mathbf{0}_{3 \times 3} & T(\phi, \theta)
\end{bmatrix},
\end{equation}
where, $\boldsymbol{J}(\boldsymbol{\eta})$ is a standard Jacobian matrix.
The REXROV2 is assumed to operate with fixed pitch and roll motion, so \eqref{model} and \eqref{jacobian} can be represented as
\begin{equation}
\begin{aligned}
    \dot{x}_v &= v_x \cos \psi - v_y \sin \psi, \\
\dot{y}_v &= v_x \sin \psi + v_y \cos \psi, \\
\dot{z}_v &= v_z, \\
\dot{\psi} &= r.
\end{aligned}
\end{equation}
\begin{figure}[t]
    \centering
         \includegraphics[width =6cm, height = 4cm]{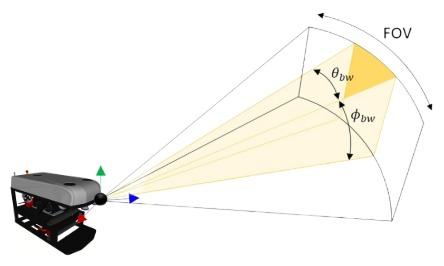}  
    \caption{Details of single beam of FLS \cite{choi2021physics}.}
    \label{ssb}
\end{figure}

\subsubsection{Sonar Model}

The sonar employed in this work is BlueView P900 NPS
multibeam sonar. The sonar plugin \cite{choi2021physics} is deployed in DAVE simulator \cite{zhang2022dave}. The sonar system operates at a frequency of 900 KHz. A single sonar beam within the FOV of the sensor is shown in Fig. \ref{ssb} which has a $90^{\circ}$ horizontal field of view ($\theta_{\text{FLS}}$), offers range $R \in [2~m,60~m]$, features a vertical beam width ($\theta_{bw}$) of $20^{o}$ and horizontal beam spacing ($\phi_{bw}$) of $0.18^{o}$. It utilizes $N_B = 512$ beams each of which is modeled
using rays.



The beams $b_i$ with range $R_i$, $i \in [1~,N_B]$ are the distances from the origin of the sonar reference frame to the first intersection between the beam and object in the FOV. The global coordinates of the obstacle points $\mathbf{p}_0$ detected by sonar in the vehicle's 2D plane can be found with
\begin{equation}
\begin{aligned}
x_{o} &= x_{v} + R \cos \theta_h , \\
y_{o} &= y_{v} + R \sin \theta_h ,
\end{aligned}
\end{equation}
where, $\theta_h \in \left[\psi - 0.5\,\theta_{\text{FLS}},\; \psi + 0.5\,\theta_{\text{FLS}}\right] $ is the horizontal azimuth angle of the reflected beam relative to the AUV heading $\psi$ in the global frame.
Further, to generate a 3D point cloud, the 2D sonar is mechanically pivoted by an elevation angle $\Theta_{\mathcal{P}}$ (measured from the horizontal plane). Then the 3D obstacle position can be represented as
\begin{equation}
\begin{aligned}
x_o &= x_{v} + R \cos \theta_h \cos \Theta_{\mathcal{P}} , \\
y_o &= y_{v} + R \sin \theta_h \cos \Theta_{\mathcal{P}} , \\
z_o &= z_{v} + R \sin \Theta_{\mathcal{P}}
\end{aligned}\label{3D_kin}
\end{equation}
By sweeping $\Theta_{\mathcal{P}}$, the sonar constructs a 3D point cloud of reflected obstacle points.

\subsubsection{Problem Statement}
Let $\mathbf{p}_v(t)^T$ be the AUV position at time $t$ and $\mathbf{p}_G$ be the goal location. Let $\mathcal{P}_o(t)$ be the set of obstacle points encountered by the vehicle at time $t$. The instantaneous distance between the vehicle and any of the obstacle be,
\begin{equation}
d(\mathbf{p}_{v}(t),\mathbf{p}_o(t))=\min_{\mathbf{p}_o\in\mathcal{P}_o}\|\mathbf{p}_{v}(t)-\mathbf{p}_o(t)\|    
\end{equation}

\begin{figure}[t]
    \centering
    \includegraphics[width = 8 cm]{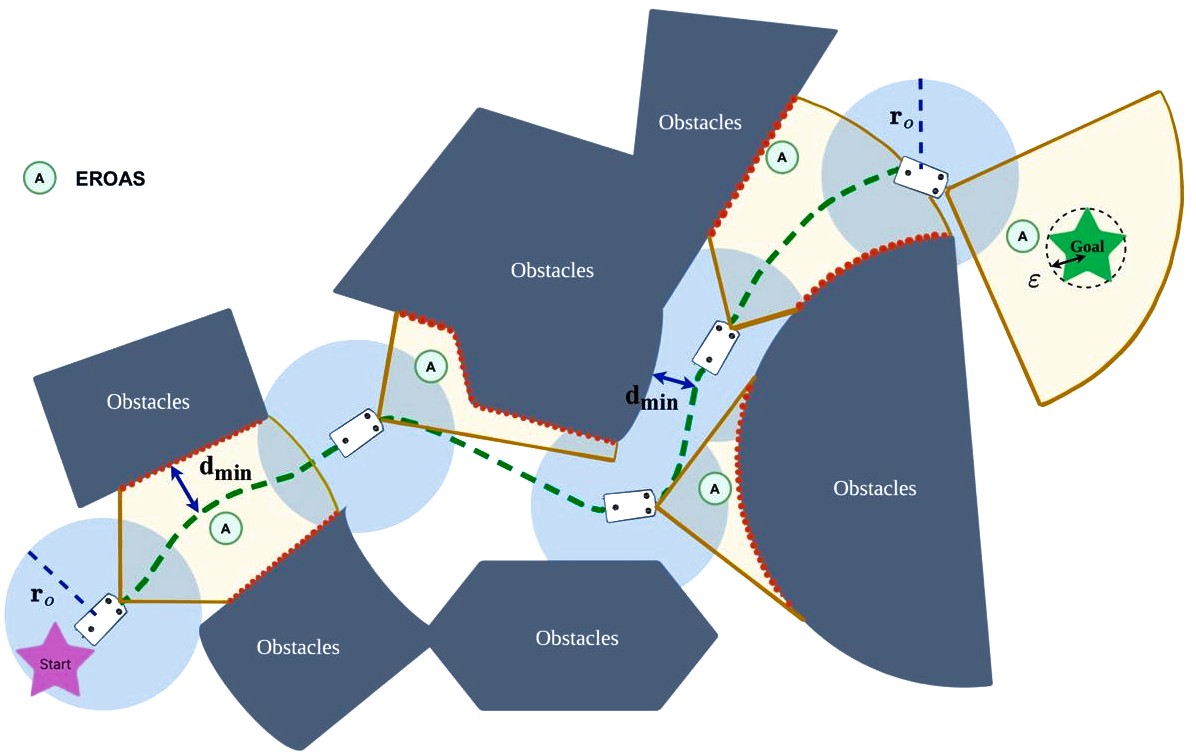}
    \caption{AUV navigating in cluttered environment}
    \label{problem_definition}
\end{figure}

The AUV needs to reach the goal within a defined tolerance, $\varepsilon$ and it should always maintain at least a distance $d_{min}$ from the obstacles to finally avoid them, as shown in Fig. \ref{problem_definition}. Hence, the following constraints
\begin{equation*}
\quad \|\mathbf{p}_{v}(T) - \mathbf{p}_G\|\le \varepsilon,\quad
d\big(\mathbf{p}_{v}(t),\mathbf{p}_o(t)\big)\ge d_{\min},\ \ \forall t\in[0,T], 
\end{equation*}
must be satisfied. Apart from this, the vehicle has limitations on actuation. There is a limit on maximum speed attained by the vehicle in all three translational directions and yaw rate. 
\begin{equation*}
    0 \leq v_x \le v_{x,\max}, |v_y|\le v_{y,\max}, \\ |v_z|\le v_{z,\max}, |r|\le r_{\max} \label{eq:act_limit}
\end{equation*}
where, $max$ represents respective maximum actuator limits. 

These actuation limits define the feasible operating limits of the AUV and directly influence maneuverability. Any control strategy must therefore generate commands that respect these bounds to avoid actuator saturation, loss of stability, or infeasible trajectories. By explicitly modeling such safety and actuation constraints, the objective of the EROAS framework is to ensure that obstacle avoidance and navigation remain achievable within the physical capabilities of the vehicle.

\subsection{Schematic of EROAS}
\label{schematic}

\begin{figure*}[t]
    \centering
    \includegraphics[width = 16 cm, height = 7cm]{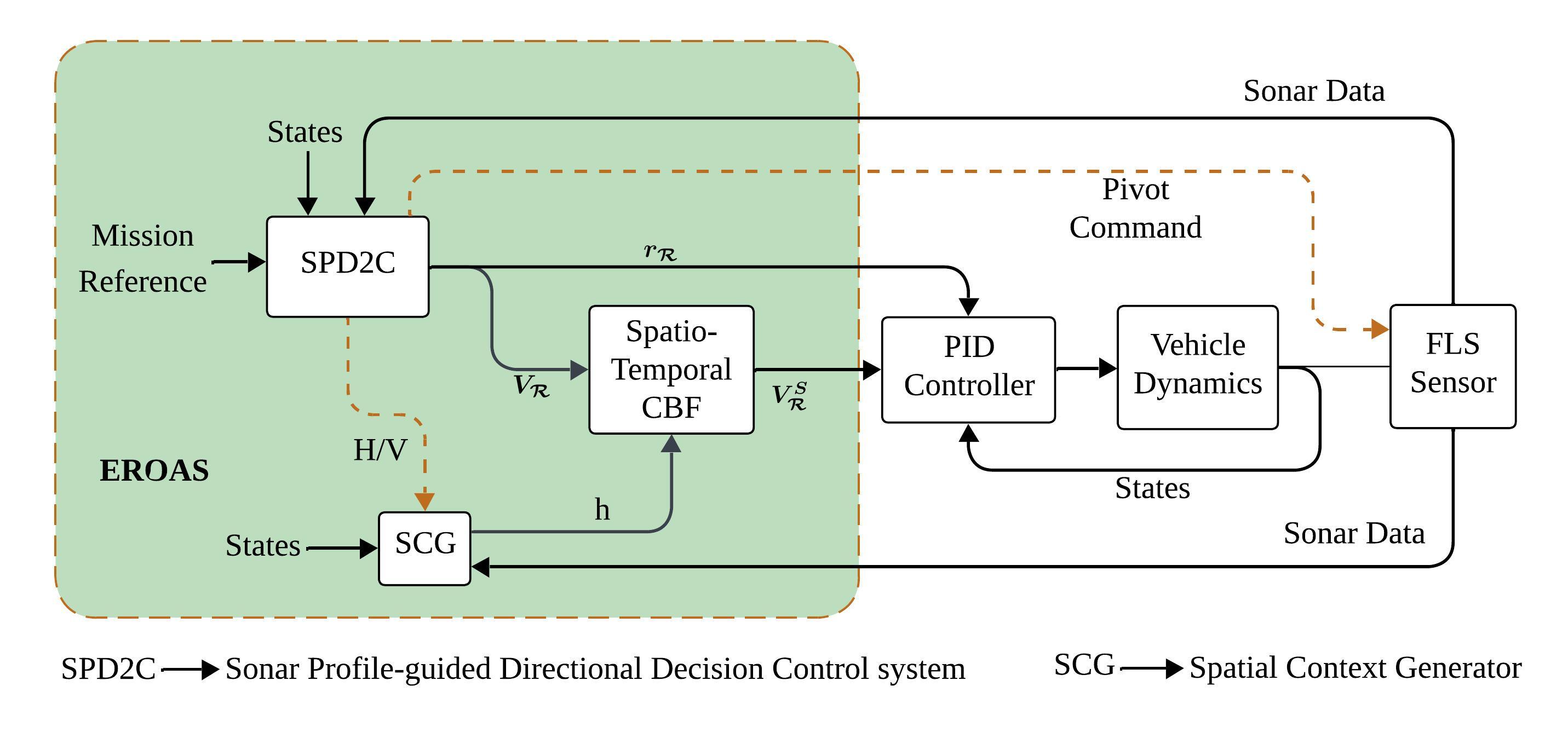}
    \caption{Schematic of EROAS framework.}
    \label{block_diagram}
\end{figure*}

This section describes the execution flow of the proposed EROAS framework, illustrated in Fig.~\ref{block_diagram}. The pipeline begins with the Sonar Profile-guided Directional Decision Control (SPD2C) system, which processes data from the 2D Forward-Looking Sonar (FLS) to generate reactive decisions. SPD2C follows a structured sequence of steps—gap finding, boundedness evaluation, convergence analysis, and adaptive pivoting of the sonar to extend 2D sensing into a 2.5D representation, thereby enabling actionable 3D situational awareness. To complement this, a Spatial Context Generator (SCG) maintains a short-horizon memory of obstacle locations, ensuring continuity of decision-making in partially observable environments. The reference commands generated by SPD2C are then refined using a Spatio-Temporal Control Barrier Function (ST-CBF), which makes use of SCG while generating safe reference outputs for speed, and heading. Finally, these safe references are tracked by a low-level PID controller, which drives the vehicle dynamics to achieve efficient and collision-free navigation. The overall strategy balances computational efficiency with safety, achieving both reduced travel time and path length while ensuring robust performance in cluttered underwater environments.


\subsection{EROAS Algorithm}
\label{eroas}
The proposed EROAS algorithm encompasses sonar data processing, obstacle detection, and local path planning strategies for real-time AUV navigation. The algorithm follows a structured approach to ensure smooth and efficient navigation with  partial observality due to  FLS and the vehicle motion. The FLS provides new measurements at every step. After receiving the sonar data the SPD2C system analyses the profile and based on certain steps and gives reference velocity and yaw rate commands. In parallel, the instantaneous sonar data is taken by SCG to create a local map of specified radius. The ST-CBF then takes in those reference velocities and SCG data and generates the safe reference velocities for the low-level controller to follow.

\subsubsection{Sonar Profile Guided Directional Decision Control (SPD2C) System}

SPD2C system is a set of sequential steps to find the reference velocities and yaw rate by analyzing the obstacle profiles. First, SPD2C identifies gaps between obstacles to determine a feasible path. If no gap is found, the system checks for boundedness, deciding whether the vehicle should turn left or right based on whether the obstacle ends in either direction. If the obstacle is unbounded in both directions, the algorithm evaluates its convergence by fitting a polynomial to the point cloud data generated by the sonar. If a feasible path is still unavailable in the 2D plane, the sonar is pivoted to acquire 3D information, allowing the vehicle to ascend or descend depending on the environment.

\begin{itemize}
    \item {\bf Gap Finding:}
Let $B=\{ b_1, b_2, \dots, b_{N_B} \}$ denote the set of all sonar beams, where $N_B$ is the total 
number of beams per scan. A beam is classified as \emph{obstacle-free} if its reflected intensity 
is strictly below a threshold $I_{\text{thr}}$ (we set $I_{\text{thr}}=15$, following \cite{xiao2024}). 
This ensures that intensities above the threshold correspond to reflections from physical objects, 
while lower values are treated as noise. The set of obstacle-free beams is defined as
\begin{equation}
    B_{\mathcal{F}} = \{\, b_i \in B \ \mid\ I_{i} < I_{\text{thr}} \,\}.
\end{equation}
Conversely, the set of beams that encounter obstacles is given by the complement: $B_{\text{o}} = B \setminus B_{\mathcal{F}}$.
We partition the obstacle-free beam set $B_{\mathcal{F}}$ into overlapping subsets of fixed cardinality, 
each representing a candidate gap. Specifically, we construct subsets of length $L=150$ consecutive beams (Here the number 150 is chosen based on the range of sonar and the size of the vehicle). 
For an index $i$ satisfying $1 \leq i \leq |B_{\mathcal{F}}|-L+1$, the $i$-th subset is defined as,
\begin{equation}
    \mathcal{G}_i = \{ b_{i}, b_{i+1}, \dots, b_{i+L-1} \}, \qquad |\mathcal{G}_i| = L.
\end{equation}
The collection of all such subsets forms,
\begin{equation}
    \mathcal{G} = \{ \mathcal{G}_1, \mathcal{G}_2, \dots, \mathcal{G}_k \}, \qquad 
    k = |B_{\mathcal{F}}|-L+1.
\end{equation}
Since each $\mathcal{G}_i$ spans $L$ consecutive beams, we identify the \emph{mid-beam} of $\mathcal{G}_i$ as
$b_{M,i} = b_{i+\lfloor L/2 \rfloor}$. The set of all mid-beams is then
\begin{equation}
    M = \{ b_{M,1}, b_{M,2}, \dots, b_{M,k} \}.
\end{equation}
To progress towards the goal, a target beam $b_{\mathcal{T}}$ is defined that corresponds to the global goal direction projected into the sonar FOV.
\begin{equation}
\mathcal{T} =
\begin{cases}
1 & \text{if } \Theta_{\mathbf{G}} < \Theta_{\mathcal{S}}, \\
\left\lfloor \dfrac{\Theta_{\mathbf{G}} - \Theta_{\mathcal{S}}}{\mathbf{r}_b} \right\rfloor & 
\text{if } \Theta_{\mathcal{S}} \leq \Theta_{\mathbf{G}} \leq \Theta_{\mathbf{E}}, \\
{N_B} & \text{if } \Theta_{\mathbf{G}} > \Theta_{\mathbf{E}},
\end{cases}
\end{equation}
where $\Theta_{\mathbf{G}}$ is the angle between the goal and the vehicle, 
$\Theta_{\mathcal{S}}$ and $\Theta_{\mathbf{E}}$ denote the angular directions of the first 
and the last sonar beams relative to the horizontal baseline. $\mathbf{r}_{b}$ is the beam width defined as,
\begin{equation}
\mathbf{r}_{b} = \frac{\Theta_{\mathbf{E}} - \Theta_{\mathbf{S}}}{N_B}.
\end{equation}
The beam chosen for navigation 
is the mid-beam in $M$ that minimizes the angular deviation from $b_{\mathcal{T}}$, 
\begin{equation}
    b_{\text{cl}} = \arg \min_{b_{M} \in M} \big| b_{M} - b_{\mathcal{T}} \big|.
\end{equation}
This ensures that among all feasible gaps, the one most closely aligned with the goal is selected for vehicle guidance.
\item {\bf Check For Boundedness}:  
If no feasible set $\mathcal{G}$ is found, it indicates the absence of a path for the AUV to 
move toward the goal. In this case, the next step is to analyze the extent of the obstacle within the 
sonar FOV. Let
$i_{\min} = \min \{ i \mid b_i \in B_{\text{o}} \}, ~
i_{\max} = \max \{ i \mid b_i \in B_{\text{o}} \}$,
denote the indices of the leftmost and rightmost beams in the obstacle set $B_{\text{o}}$, respectively. 
Let $1$ and $N_B$ denote the indices of the first and last beams of the sonar FOV.  

The classification is then defined as
\begin{equation}
\mathcal{O} =
\begin{cases}
\text{BO}   & \text{if } 1 < i_{\min} \ \text{and}\ i_{\max} < N_B, \\
\text{LUBO} & \text{if } i_{\min} = 1 \ \text{and}\ i_{\max} < N_B, \\
\text{RUBO} & \text{if } 1 < i_{\min} \ \text{and}\ i_{\max} = N_B, \\
\text{UBO}  & \text{if } i_{\min} = 1 \ \text{and}\ i_{\max} = N_B.
\end{cases}
\end{equation}
The navigation strategy follows directly from this classification: For a BO, the AUV turns toward the side containing the goal. For a LUBO, the AUV turns right. For a RUBO, the AUV turns left. For a UBO, the algorithm proceeds to the next step (convergence check).


\item {\bf Check for Convergence}:  
If the obstacle is unbounded on both sides (UBO), the next step is to evaluate whether its geometry is 
converging (convex) or diverging (concave/wall-like). For this, the obstacle beams $B_{\text{o}}$ are first 
transformed from polar sonar coordinates to global Cartesian coordinates
We approximate the obstacle boundary in the $(x,y)$-plane by fitting a quadratic polynomial $f(x) = ax^2 + bx + c$,
to the points in $C_{\text{o}}$ using least-squares regression. The coefficient $a$ encodes the 
curvature of the parabola and thus characterizes the local convexity of the obstacle and is classified as
\begin{equation}
\mathcal{O}_{\text{conv}} =
\begin{cases}
\text{Convex (Converging)} & \text{if } a \geq C_{\text{th}}, \\
\text{Concave/Wall (Diverging)} & \text{if } a < C_{\text{th}}.
\end{cases}
\end{equation}
The navigation policy is defined as follows: If $\mathcal{O}_{\text{conv}}=$ Convex, the AUV turns toward the goal-side slope of the fitted 
    parabola and goes to the gap-finding step. This ensures that the vehicle exploits the 
    narrowing geometry to progress toward the target. If $\mathcal{O}_{\text{conv}}=$ Concave/Wall, the geometry indicates that no feasible passage 
    exists in current plane. In this case, algorithm proceeds to sonar pivoting 
    for vertical search. 
This convexity-based classification prevents the vehicle from being trapped in local minima 
caused by concave boundaries or walls.


\item{\bf Pivoting the Sonar}: When no feasible 2D solution exists, the AUV engages a vertical pivot scan of the FLS
to acquire short-horizon vertical information. Let $\Theta_{\mathcal{P},i} \in \mathbb{Z} $ denote a pivot angle of the sonar,
\begin{equation}
\mathcal{P} = \{\Theta_{\mathcal{P},1},\Theta_{\mathcal{P},2},\dots,\Theta_{\mathcal{P},N_P}\}, 
\end{equation}
be the set of pivot angles swept by the sensor. At a pivot angle $\Theta_{\mathcal{P}}$ the FLS returns a beam-wise classification; denote by
\begin{equation}
B_{\mathcal{F}}(\Theta_{\mathcal{P},i})\subseteq B, \forall i \in [1,~N_P].
\end{equation}
The set of indices of obstacle-free beams observed at that pivot angle. To focus on the central portion of the sonar frustum, we define the central beam index sector as,
\begin{equation}
\mathcal{I}_c = \{i\in\mathbb{Z}\mid i_{\min}^{(c)} \le i \le i_{\max}^{(c)}\},
\end{equation}
where in our implementation $i_{\min}^{(c)}=100$ and $i_{\max}^{(c)}=400$.
Further, a pivot angle $\Theta_{\mathcal{P}}$ is considered acceptable if the corresponding scan has all the beams in central sector as free beams. We define the accepted-pivot set,
\begin{equation}\label{eq:B_pivot}
\mathcal{P}_{\text{acc}} \;=\; \big\{ \Theta_{\mathcal{P},i}\in\mathcal{P} \; |B_{\mathcal{F}}(\Theta_{\mathcal{P},i})\cap\mathcal{I}_c\,\big \}.
\end{equation}
From $\mathcal{P}_{\text{acc}}$, we create groups of consecutive acceptable pivot angles. we extract groups of maximal runs of consecutive pivot angles and then form fixed-length candidate groups. Let $L$ denote the group length (we use $L=30$). 
Further, constructing the collection of length-$L$ consecutive groups as,
\begin{equation}
\mathcal{S}_a \!\!=\!\! \left\{ S_j \!\mid \! S_j \!= \!\Theta_{\mathcal{P},j},\dots,\Theta_{\mathcal{P},j+L-1}\}, S_j\subseteq\mathcal{P}_{\text{acc}} \right\}.
\end{equation}
Indices here refer to the ordered sequence of accepted pivot angles; only groups fully contained in $\mathcal{P}_{\text{acc}}$ are included. For each candidate group $S_j\in\mathcal{S}_a$ define its midpoint angle (arithmetic mean of end angles)
\begin{equation}
\Theta_{\text{mid},j} \;=\; \frac{\Theta_{\mathcal{P},j} + \Theta_{\mathcal{P}, j+L-1}}{2}.
\end{equation}
Collect the mid-angles into the set
\[
\Theta_{\text{mid}} \;=\; \{ \Theta_{\text{mid},j} \mid S_j\in\mathcal{S}_a \}.
\]

Let $\Theta_{\mathcal{T}'}$ be the goal direction expressed as an elevation angle in the same pivot domain.

\begin{equation}
\Theta_{\mathcal{T}'} =
\begin{cases}
\Theta_{\mathcal{P},1} &\!\!\! \text{if } \Theta_{\mathbf{G_v}} < \Theta_{\mathcal{P},1}, \\
\!\!\Theta_{\mathbf{G_v}} - \Theta_{\mathcal{P},1} & 
\!\!\!\text{if } \!\Theta_{\mathcal{P},1} \!\leq\! \Theta_{\mathbf{G_v}} \!\leq \!\Theta_{\mathcal{P},N_P}, \\
\Theta_{\mathcal{P},N_P} & \!\!\!\text{if } \Theta_{\mathbf{G_v}} > \Theta_{\mathcal{P},N_P},
\end{cases}
\end{equation}
where $\Theta_{\mathbf{G_v}}$ is the angle between the goal and the vehicle in vertical direction, 
$\Theta_{\mathcal{P},1}$ and $\Theta_{\mathcal{P},N_P}$ denote the first and last sonar pivot angles from the horizontal plane. The selected pivot mid-angle is the element of $\Theta_{\text{mid}}$ closest to the goal,
\begin{equation}\label{eq:Theta_cl}
\Theta_{\text{cl}} \;=\; \arg\min_{\Theta\in\Theta_{\text{mid}}} \; \big| \Theta - \Theta_{\mathcal{T}} \big|.
\end{equation}
This ensures that among all feasible gaps, the one most closely aligned with the goal is selected for vehicle guidance. If $\Theta_{\text{mid}}=\varnothing$ then no acceptable vertical gap is found and the algorithm falls back to the 2D policy (turn and re-evaluate).

Finally, to navigate through the required horizontal and vertical gap, the AUV follows a velocity mapping approach, where the forward velocity \( v_{x,\mathcal{R}} \) and vertical velocity \( v_{z,\mathcal{R}} \) are adjusted accordingly.
\begin{eqnarray}
    v_{x,\mathcal{R}}& = & K_v (\psi_{\text{max}} - |\psi_{\mathcal{R}}|), \\
    v_{z,\mathcal{R}} & = & v_{x,\mathcal{R}} \times \tan \left( {\Theta_{\text{cl}}} \right),\\
    r_{\mathcal{R}} & = & K_t * \psi_{\mathcal{R}},
\end{eqnarray}
with $K_v$, $K_t$ and $\psi_{\text{max}} = \theta_{FLS}$  are constants and 
\begin{equation}
   \psi_{\mathcal{R}} = \psi(b_{cl}) = \frac{\pi}{2} - \left(K_r*b_{cl} + \frac{\pi}{4} \right) \label{yawangle},
\end{equation} 
where, $ K_r $ is a tuned gain.
The SPD2C system gives zero lateral velocity $v_{y,\mathcal{R}}$ as reference to ST-CBF.
Thus the AUV adjusts its motion vertically to bypass obstacles and continue toward the goal. 
If no feasible path exists in the vertical plane, it initiates a leftward turn while re-evaluating the 
environment until a solution emerges. However, due to the sonar’s limited FOV relying only on 
local sensing may still risk collisions. To overcome this, the SCG maintains a 
short-term memory of obstacle points, while the ST-CBF enforces 
safety constraints over reference commands $v_{x,\mathcal{R}}$, $v_{y,\mathcal{R}}$, and $v_{z,\mathcal{R}}$. This integration extends situational awareness 
and ensures safer navigation under partial observability.
\end{itemize}

\subsubsection{Spatial Context Generator (SCG)}
\label{SCG}
\begin{figure}[b]
    \centering
    \includegraphics[width = 8.5cm, height = 6.5 cm]{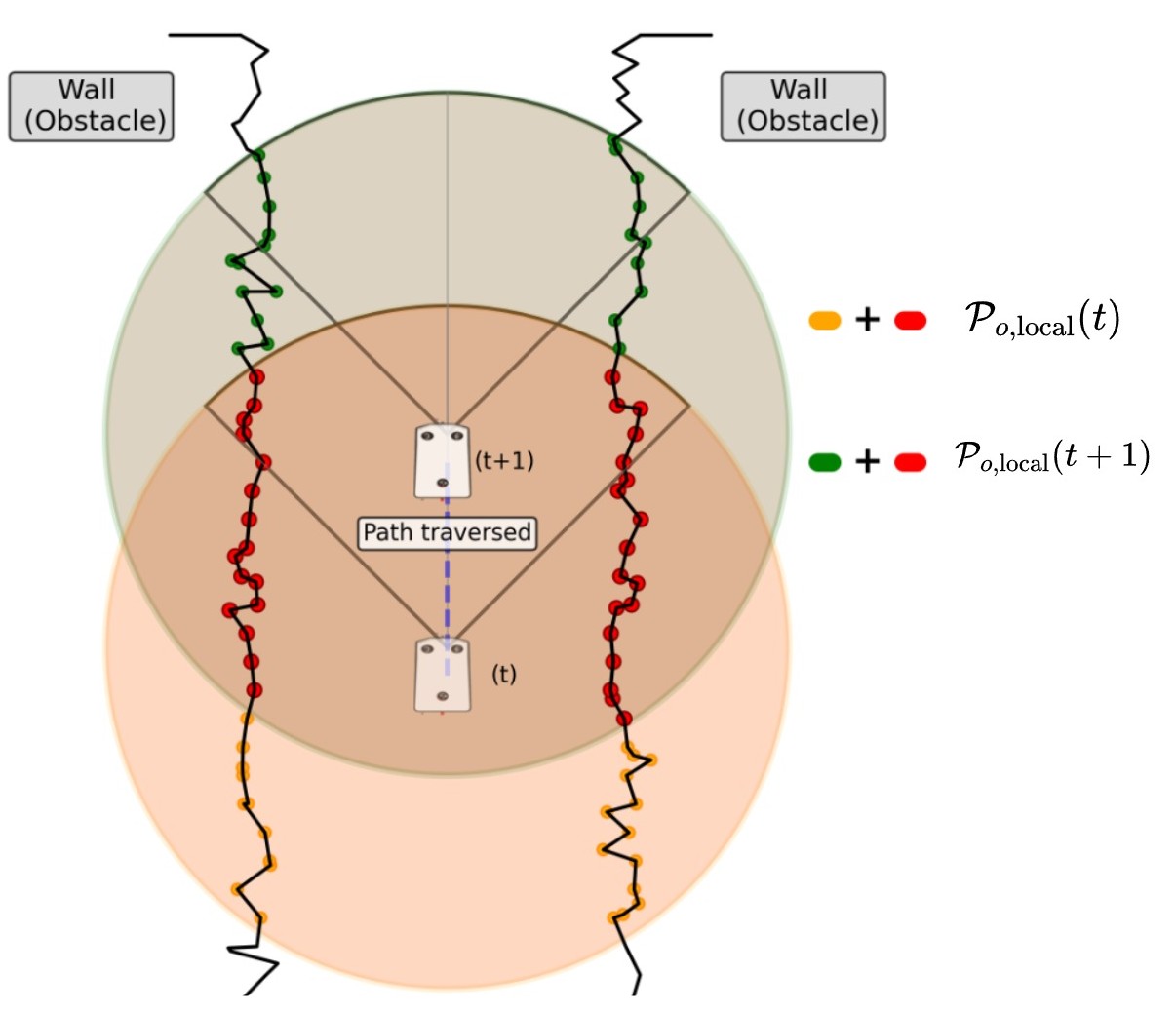}
    \caption{Spatial Context Generator}
    \label{context_map}
\end{figure}

This block builds and maintains a short-horizon map of obstacles using FLS measurements, allowing the system to retain information about previously encountered obstacles—an essential capability not only for anticipating future scenarios but also for ensuring continuous safe navigation. At each control cycle $t$, the sonar returns a 
set of obstacle points $\mathcal{P}_o(t)=\{\mathbf{p}_{o,i}(t)\}_{i=1}^{N(t)}$. To ensure robustness under
partial observability, the SCG maintains a memory of points within a radius-$\mathbf {r_o}$ ball centered on
the vehicle position:
\begin{equation}
\mathcal{P}_{o,\text{local}}(t) = \left\{\mathbf{p_o}\in\mathbb{R}^3 \ \big|\ \|\mathbf{p_o}-\mathbf{p}_v(t)\|_2 \leq \mathbf{r}_o \right\}.
\end{equation}
This memory is updated by retaining points that remain within the radius while discarding those that move outside, and adding the new detections from the current FLS scan.
The update rule is
\begin{equation*}
\mathcal{P}_{o,\text{local}}(t+1) = 
\;\Big\{\mathbf{p}_o \in \mathcal{P}_{o,\text{local}}(t) \;\big|\; 
\|\mathbf{p}_o-\mathbf{p}_v(t^+)\|_2 \le \mathbf{r}_o \Big\} 
\end{equation*}
\begin{equation}
\;\cup \Big( \mathcal{P}_o(t+1) \cap \mathbb{B}_{\mathbf{r}_o}(\mathbf{p}_v(t+1)) \Big).
\end{equation}
This is illustrated in Fig.~\ref{context_map}. 
From $\mathcal{P}_{o,\text{local}}(t)$, the SCG identifies the closest obstacle point:
\begin{equation}
\mathbf{p}_{o,\text{cl}}(t) = \arg\min_{\mathbf{p_o}\in \mathcal{P}_{o,\text{local}}(t)} \|\mathbf{p_o}-\mathbf{p}_v(t)\|_2.
\end{equation}
Depending on the SPD2C output, the SCG provides context in two different ways: \textbf{Horizontal (H) case:} When SPD2C selects a horizontal maneuver, the SCG collects 
    obstacle points $\mathbf{p}_{o,\text{cl}}^{XY}=[x_{o,\text{cl}},y_{o,\text{cl}}]^\top$. \textbf{Vertical (V) case:} When SPD2C selects a vertical maneuver, the sonar pivoting provides 
These obstacle points $\mathbf{p}_{o,\text{cl}}^{XY}$ and $\mathbf{p}_{o,\text{cl}}^{XZ}$ are fed to ST-CBF to ensure the 
closest relevant obstacle is always considered for either horizontal or vertical avoidance. This dual-context mechanism is essential for generating safe references in both planes under cluttered and partially observable conditions.

\subsubsection{Spatio-Temporal Control Barrier Function (ST-CBF)}
The well-established framework of CBFs is utilised to effectively resolve conflicts in obstacle avoidance and navigation. With CBF \cite{ames2019control}, a safe set is defined for the states of autonomous vehicles. Staying inside these safety regions guarantees that there will be no conflict or collisions for the AUVs. 
Let $\mathbf{R}_o>0$ denote a conservative obstacle radius. The CBF is defined as
\begin{equation}
h(\mathbf{p}_v,t)=\|\mathbf{p}_v(t)-\mathbf{p}_{o, \text{cl}}(t)\|_2^2 - \mathbf{R}_o^2 ,
\end{equation}
where $\mathbf{p}_v$ is the vehicle position and $\mathbf{p}_{o,\text{cl}}$ is the closest obstacle point from the SCG. The safe set is then
\begin{equation}
\mathcal{C}(t)=\{\mathbf{p}_v \mid h(\mathbf{p}_v,t)\ge 0\}.
\end{equation}
Let $V=[v_x\ v_y\ v_z]^\top$ denote the vehicle velocity and 
$V_{\mathcal{R}}=[v_{x,\mathcal{R}}\ v_{y,\mathcal{R}}\ v_{z,\mathcal{R}}]^\top$ the reference velocity proposed by the SPD2C module. 
The safe control action is obtained from the QP (quadratic programming),
\begin{equation}
\min_{V}\ \tfrac{1}{2}\|V - V_{\mathcal{R}}\|^2 ,
\end{equation}
subject to
\begin{equation}
\dot{h} \geq -\alpha(h).
\end{equation}
A linear class-$\mathcal{K}$ functions is employed for $\alpha(h)=k h$, with $k>0$.
Expanding the cost function,
\begin{equation}
\|V \!- \!V_R\|^2\! =\! (v_x - v_{x,R})^2 + a(v_y - v_{y,R})^2 + b(v_z - v_{z,R})^2 ,
\end{equation}
where $a,b \in \{0,1\}$ act as selectors for the active plane. Specifically, when SPD2C outputs 
a horizontal maneuver ($H$), we set $a=1, b=0$ and solve the QP in the $XY$-plane; when it outputs 
a vertical maneuver ($V$), we set $a=0, b=1$ and solve in the $XZ$-plane. The resulting safe reference 
velocity is denoted $V_{\mathcal{R}}^S=[v^S_{x,\mathcal{R}}\ v^S_{y, \mathcal{R}}\ v^S_{z,\mathcal{R}}]^\top$ fed to the PID controller that enables AUV to avoid obstacles in the uncertain cluttered environment. 
The barrier depends not only on the instantaneous FLS measurements but also on the time-varying memory 
$\mathcal{P}_{o,\text{local}}(t)$ maintained by the SCG. Thus, even when an obstacle leaves the sonar 
field of view due to occlusion or limited range, it remains in memory until it exits the predefined 
radius-$\mathbf{r}_o$ ball. This spatio-temporal coupling ensures that $h$ continues to provide safety constraints, 
enabling robust navigation under partial observability. The details of the EROAS are presented in Algorithm  \ref{algo}.

\begin{algorithm}[t]
\caption{EROAS Loop with SPD2C, SCG, and ST\mbox{-}CBF}
\begin{algorithmic}[1]
\Require sonar scan $I$, mission reference $(G)$, current state, system parameters
\Ensure safe reference commands $(V^{S}_{\mathcal{R}},\,r_{\mathcal{R}})$

\While{goal not reached}
  \State $I \gets \text{Read FLS data}()$ \Comment{obtain the latest sonar scan}
  \State $(V_{\mathcal{R}},\, r_{\mathcal{R}},\, \mathrm{mode},\, \Theta_{\mathcal{P}}) \gets \mathrm{SPD2C}(I,\text{state},g)$
  \If{$\Theta_{\mathcal{P}}$ is defined}
    \State \text{Pivot the FLS}$(\Theta_{\mathcal{P}})$ \Comment{command a pivot scan if needed}
  \EndIf
  \State $(h,\text{context}) \gets \mathrm{SCG}(I,\text{state},\mathrm{mode})$ \Comment{update local obstacle memory}
  \State $V^{S}_{\mathcal{R}} \gets \mathrm{ST\mbox{-}CBF}(V_{\mathcal{R}},h)$ \Comment{apply CBF filter to nominal references}
  \State $r_{\mathcal{R}} \gets \mathrm{clip}\bigl(r_{\mathcal{R}},[-r_{\max},r_{\max}]\bigr)$ \Comment{enforce yaw-rate limits}
  \State \text{Apply} $(V^{S}_{\mathcal{R}},\, r_{\mathcal{R}})$ \text{to the low-level controller and}
  \text{update state}
\EndWhile
\end{algorithmic}
\label{algo}
\end{algorithm}



In the next section, simulations and edge-device tests that evaluate each step of this pipeline are presented and quantify its impact on safety and efficiency.

\section{Performance Evaluation} \label{result}
In this section, we assess the performance of EROAS across diverse underwater scenarios, including 
horizontal and vertical avoidance in cluttered 3D environments. The evaluation highlights the roles 
of SPD2C, SCG, and ST-CBF in ensuring safe navigation under partial observability and compares EROAS 
with baseline methods DWA and APF in terms of trajectory efficiency, smoothness, and safety. 
Finally, hardware-in-the-loop (HIL) experiments confirm the practical applicability of the framework 
for real-world AUV operations.

\subsection{Scenario Setup}

As discussed in previous sections, the DAVE simulator is used to test the proposed algorithm. A Gazebo environment with various complex-shaped static obstacles is set up to evaluate the algorithm's robustness across all scenarios. Simulations were run on an NVIDIA GeForce GTX 1050 Ti GPU with Driver Version 550.54.14 and CUDA Version 12.4. The DAVE simulator was deployed using ROS 1 (Noetic) on an Ubuntu 20.04 system, alongside deployment of the algorithm on edge devices.
The results for 2D and 3D avoidance are discussed further in following sections. The parameter values used in the Gazebo simulation are presented in Table \ref{tab:parameters}. These values were determined through a heuristic approach to optimize performance for effective obstacle avoidance.

\begin{figure}[t]
    \centering
    \includegraphics[width = 8.7 cm, height = 7cm]{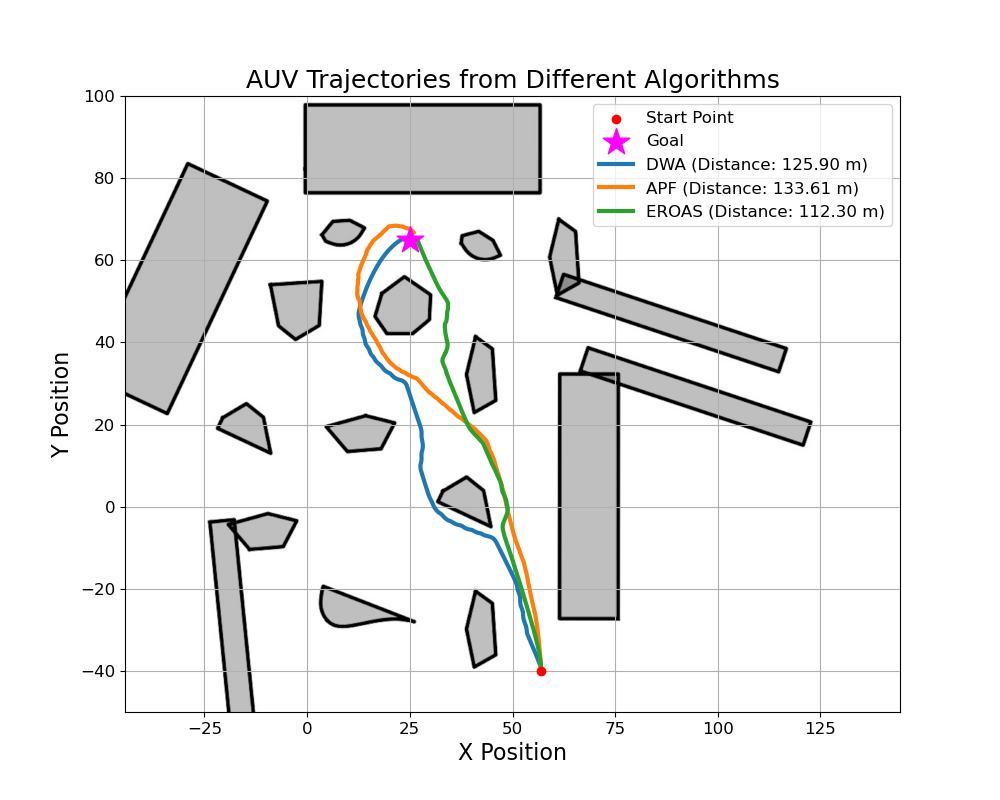}
    \caption{Top view of AUV navigating through a 3D environment in Gazebo.}
    \label{2d}
\end{figure}

\begin{table}
\caption{Parameter values used in the given simulation results}
    \centering
    \renewcommand{\arraystretch}{1.25} 
    \setlength{\tabcolsep}{8pt} 
    \begin{tabular}{|c|c|c|}
        \hline
        \textbf{Sr No} & \textbf{Parameter} & \textbf{Value} \\ 
        \hline
        1 & $K_t$ ($1/sec$) & 0.12 \\ 
        \hline
        2 & $K_v$ ($m/sec-rad$) & 0.35 \\ 
        \hline 
        3 & $K_r$$(rad)$ & 0.175 \\
        \hline 
        4 & Convexity threshold $a$ & 0.02 \\ 
        \hline
        5 & CBF memory radius (meters)  & 15 \\ 
        \hline
    \end{tabular}    
    \label{tab:parameters}
\end{table}

\begin{figure}[b]
    \centering
    \begin{tabular}{cc}
      \includegraphics[width =8.3cm]{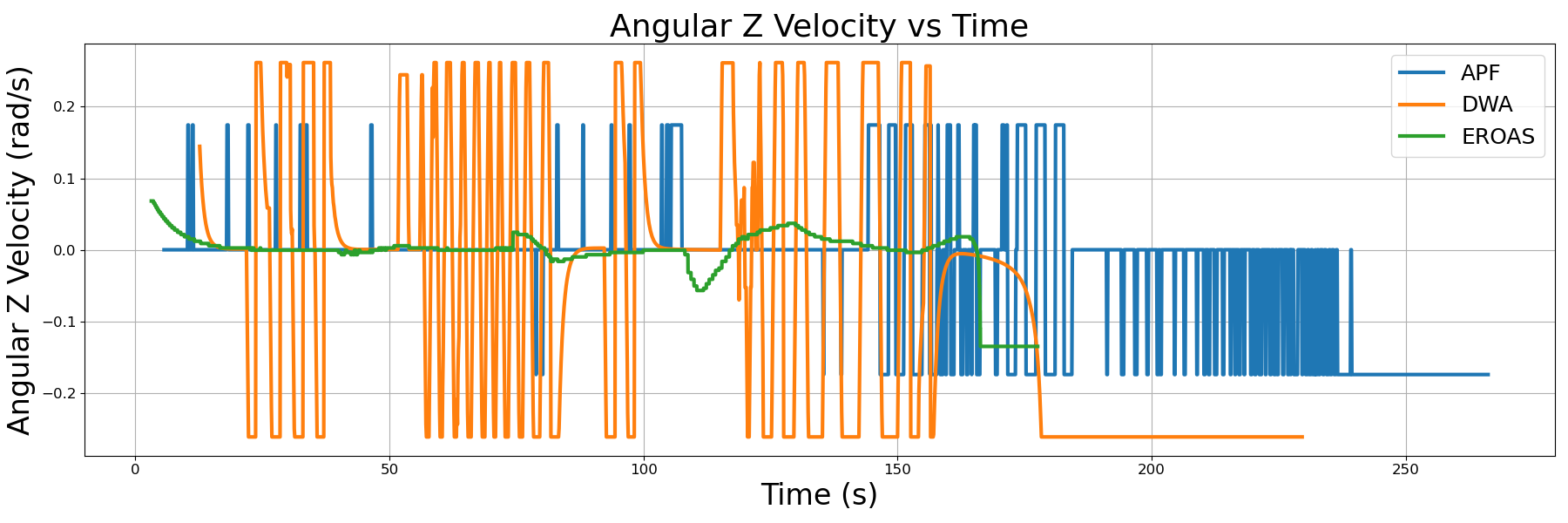}  
    \end{tabular}
    \caption{Yaw velocity with different algorithms.} 
    \label{t1}
\end{figure}

\subsection{Baseline Algorithms}
To provide a benchmark for evaluating EROAS, we compare it against two widely used motion planning approaches: APF \cite{Zhongxian2023} and DWA \cite{Juan2025} \cite{Jiahao2024}. These methods are well-established in the field of robot motion planning. APF relying on attractive and repulsive forces to guide navigation, while DWA focuses on generating dynamically feasible trajectories within the robot’s velocity space. Including these baselines allows for a comprehensive evaluation of the proposed method’s performance in terms of path length, time of flight, and command output smoothness.

\begin{figure}[t]
    \centering
    \includegraphics[width = 8.7 cm, height = 7 cm]{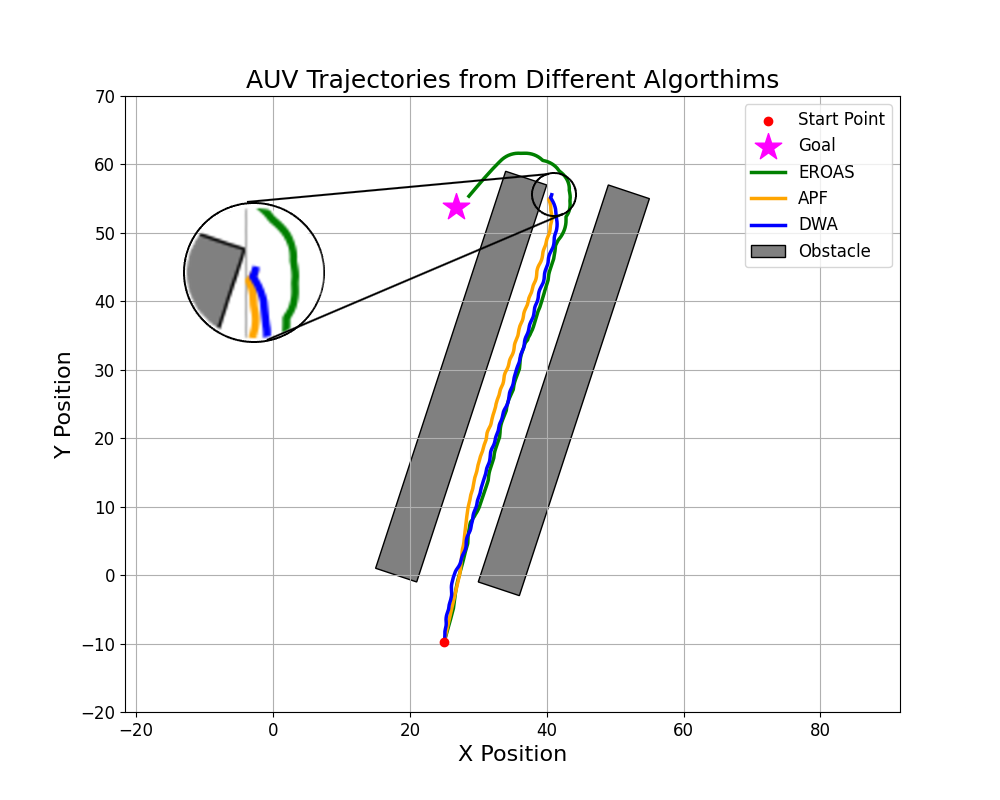}
    \caption{Planar navigation comparison in cluttered corridor}
    \label{dyn_zoom}
\end{figure}

\begin{figure*}[t]
    \centering
    \begin{tabular}{cc}
      \includegraphics[width = 8cm, height = 5cm]{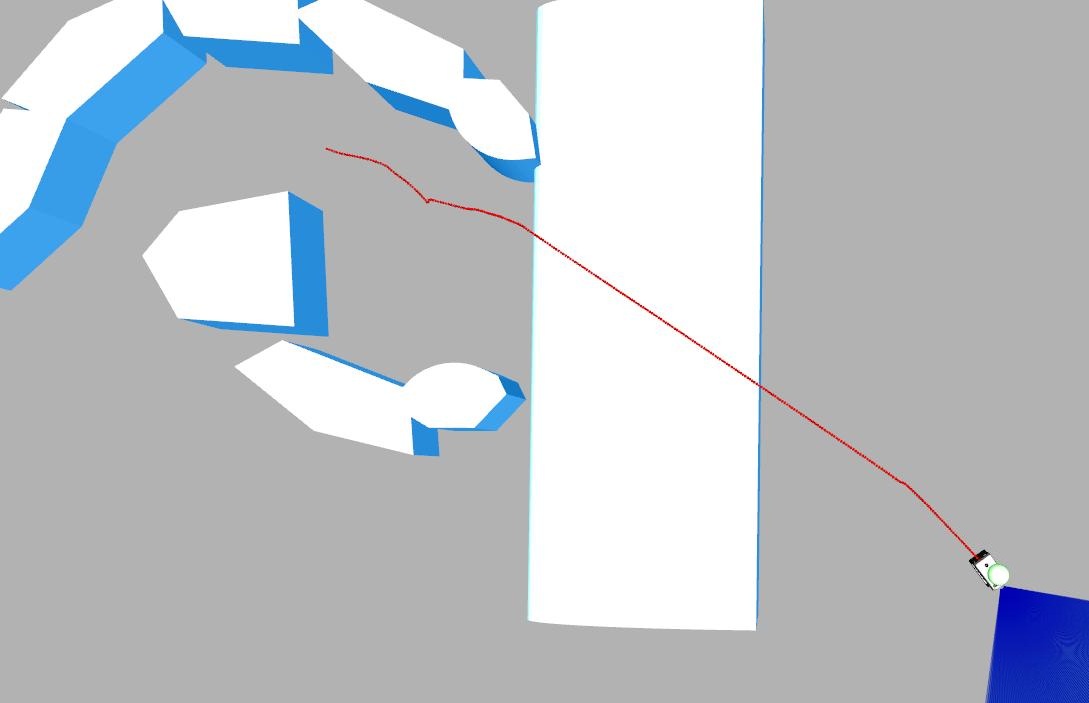}    &  \includegraphics[width = 8cm, height = 5cm]{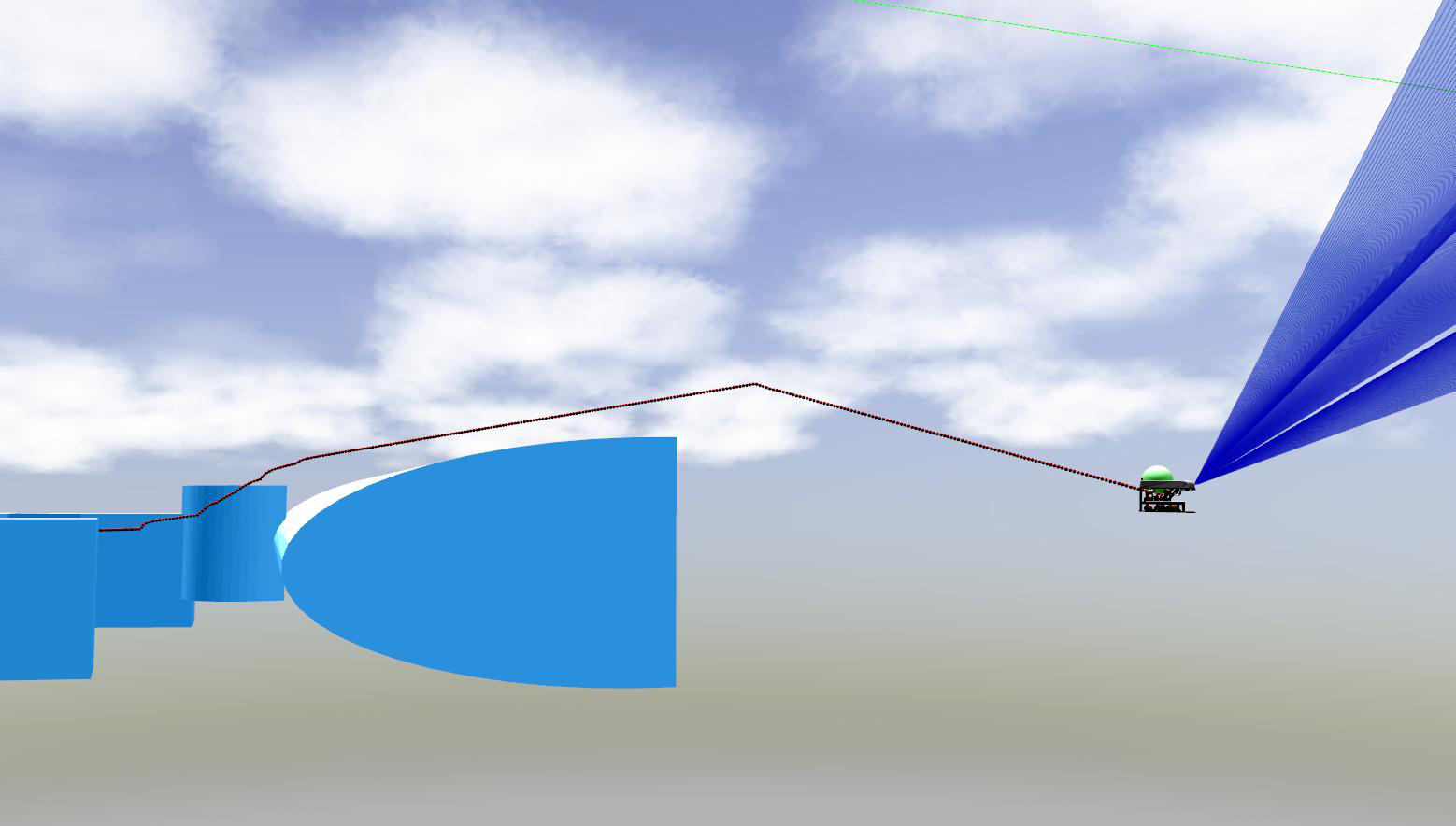} \\
        (a) & (b) 
    \end{tabular}
    \caption{(a)~Top view of AUV avoiding obstacle in 3D Gazebo environment and b)~Side view of AUV avoiding obstacle in 3D Gazebo environment.} 
    \label{topview}
\end{figure*}


\subsection{Comparative Study}
The effectiveness of the proposed avoidance algorithm was evaluated by navigating the AUV through a complex uncertain environment with multiple obstacles. As illustrated in Fig. \ref{2d}, the red dot denotes the start position, while the magenta star indicates the goal location. Although all three algorithms successfully guide the AUV to the target location, the proposed EROAS algorithm demonstrates superior performance by achieving the shortest and fastest trajectory. Specifically, the total trajectory lengths for EROAS, DWA, and APF were measured as $112.30~\text{m}$, $125.9~\text{m}$, and $133.61~\text{m}$, respectively. The corresponding times taken to complete these trajectories were $190$~s for EROAS, $220$~s for DWA, and $260$~s for APF.   For fair comparison between the algorithms, the max vehicle velocity was set to $1~\text{m/s}$ and max angular rate to $15~\text{deg/s}$.
This efficiency contributes to an extended operational range for the AUV and highlights the effectiveness of the proposed algorithm in complex navigation scenarios. In addition, Fig. \ref{t1} shows that EROAS exhibits lower yaw-rate amplitudes and fewer oscillations over time compared to others, indicating smoother steering, reduced actuation effort and potentially lower mechanical wear on the propulsion system. This is further supported by the average angular jerk values observed in simulation: DWA yielded $1.02~\text{rad/s}^3$, APF $2.42~\text{rad/s}^3$, while EROAS achieved the lowest at $0.42~\text{rad/s}^3$, indicating smoother and more stable control. This corresponds to a $58.8\%$ reduction compared to DWA and an $82.4\%$ reduction compared to APF, highlighting the superior smoothness and stability of EROAS during navigation.

Also, as discussed in Section \ref{SCG}, the importance of a SCG is highlighted in  Fig. \ref{dyn_zoom} in which an AUV must navigate a narrow tunnel-like structure and execute a sharp turn at the end to reach its goal. Traditional methods like APF and DWA fail in this scenario because they depend only on the current sensor FOV and lack memory of previously detected obstacles. In contrast, SCG in EROAS retains past information while incorporating current obstacle data, enabling the proposed method to successfully reach the goal and avoid corner obstacles even when they are outside the sensor’s FOV.

\subsection{Navigation in 3D With 2.5D Sonar}

Further, the performance of EROAS is evaluated in a scenario with non-convex obstacles. These obstacles were successfully detected and avoided by the vertical case of SPD2C-SCG mudule of the EROAS algorithm which activated the 3D avoidance mode as illustrated in the Fig.~\ref{topview}. It can be seen that the AUV ascended to avoid collisions and then descends back to resume its path towards the goal. This is illustrated in two views: the top view of the avoidance maneuver (Fig. \ref{topview} (a)), while the side view is shown in Fig. \ref{topview} (b). In these figures, the green sphere represents the goal. For clarity in the figures, the Gazebo water texture was turned off during recording.

\subsection{Hardware-in-Loop Simulation}
\begingroup
\renewcommand{\arraystretch}{1.3} 
\begin{table}[h]
\centering
\caption{Performance of the EROAS algorithm on different edge devices.}
\label{tab:edge_performance}
\begin{tabular}{lccc}
\hline
\textbf{Board} & \textbf{Frame Rate} & \textbf{Power Consumption} & \textbf{CPU Usage} \\
\hline
Jetson Orin NX   & 8\,Hz & 5.5\,W (idle: 2.5\,W) & $\sim$20\% \\
Jetson Xavier NX & 8\,Hz & 3.6\,W (idle: 2.3\,W) & $\sim$80\% \\
\hline
\end{tabular}
\end{table}
\endgroup

In addition to validating our algorithm in a simulation environment, we implemented it on two edge computing platforms commonly used in AUVs: the NVIDIA Jetson Orin NX and Xavier NX. The former offers a peak performance of 100 TOPS with 16 GB of RAM, and the latter provides 21 TOPS with 16 GB of RAM. Both devices successfully executed the EROAS framework within the desired operational constraints. During testing, the EROAS output frequency was limited to 8 Hz, which is higher than the 4 Hz sampling rate of the used onboard FLS sensor. This ensures that the algorithm operates in real-time and remains synchronized with the sensor inputs, without introducing latency or processing backlogs. The evaluation was conducted using a master–slave ROS Noetic configuration, where the host PC handled the simulation, while the EROAS nodes were deployed and executed on the edge device. The detailed performance metrics for each device are presented in Table~\ref{tab:edge_performance}. These results support the claim that EROAS is lightweight algorithm for embedded deployment while maintaining real-time performance.

\section{Conclusion} \label{conclusion}

In this work, we introduced EROAS, an efficient reactive obstacle avoidance strategy for AUVs operating in cluttered 3D environments. The framework integrates three core components: SPD2C, which enables rapid gap detection and reference command generation in both horizontal and vertical planes; the SCG, which extends situational awareness by maintaining short-term obstacle memory under partial observability; and the ST-CBF, which enforces safety through spatio-temporal constraints on the generated references. A distinctive feature of EROAS is its use of a pivoting 2D FLS to achieve 2.5D perception, providing targeted vertical awareness without the cost or complexity of full 3D sensing. EROAS was validated through simulations, edge-device experiments, and hardware-in-the-loop tests, demonstrating its practical feasibility. With EROAS, the AUV shows 10.8\% and 15.9\% reduction in path length traversed, compared to DWA and APF, respectively. Additionally, there are 13.6\% and 27\% reduction in travel time compared to DWA and APF, respectively. EROAS also demonstrates smoother steering and lower actuation effort. Overall, EROAS delivers a balanced solution that combines efficiency, safety, and adaptability.

The system has two main limitations. First, the locally reactive nature of CBFs, may yield suboptimal navigation in complex spaces with dynamic obstacles. In addition, azimuth-induced depth errors are inherent to FLS, limiting the precision of navigation. Future work will address these by integrating global path planners, real-time SLAM for mapping, and motion-predictive CBFs for dynamic obstacles.


\balance


\end{document}